\documentclass{article}

\usepackage{arxiv}

\usepackage[utf8]{inputenc}
\usepackage[T1]{fontenc}
\usepackage{amsmath,amssymb,amsfonts}
\usepackage{booktabs}
\usepackage{array}
\usepackage{multirow}
\usepackage{graphicx}
\usepackage{nicefrac}
\usepackage{microtype}
\usepackage{url}
\usepackage[hidelinks]{hyperref}
\usepackage{float}
\usepackage{placeins}

\newcommand{\secondbest}[1]{\underline{#1}}
\graphicspath{ {./images/} }

\title{LAST-RAG: Literature-Anchored Stochastic Trajectory Retrieval-Augmented Gemeration for Knowledge-Conditioned Degradation Model Selection}

\author{
 Hanbyeol Park \\
  Department of Industrial Engineering\\
  Pusan National University\\
  Busan, Republic of Korea\\
  \texttt{pb104@pusan.ac.kr} \\
  \And
 Hyerim Bae \\
  Department of Industrial Engineering\\
  Pusan National University\\
  Busan, Republic of Korea\\
  \texttt{hrbae@pusan.ac.kr} \\
}

\begin{document}
\maketitle
\begin{abstract}
Stochastic-process-based degradation modeling is a core approach for estimating the distribution of remaining useful life (RUL); however, the selection of an appropriate stochastic process has not been sufficiently addressed. Existing model selection methods mainly rely on the statistical fit of the observed health indicator (HI) trajectory, but this approach may select a model that is inconsistent with the underlying degradation mechanism when the observation window is short or the signal is highly noisy. To address this issue, this paper proposes Literature-Anchored Stochastic Trajectory Retrieval-Augmented Generation (LAST-RAG). The proposed method uses both the observed HI trajectory and domain-specific context, and hierarchically conditions the candidate degradation model space based on theoretical and mechanical evidence retrieved from a local evidence bank. In addition, Rule-based Confidence Reasoning with Uncertain State (RCRUS) is introduced to prevent candidate models from being prematurely eliminated when hierarchical decisions are uncertain. Simulation-based experiments demonstrate that the proposed method outperforms statistical, prognostic, and uncertainty-aware baselines in both Wiener/gamma family classification and detailed degradation model classification. Ultimately, this study reframes degradation model selection from a purely statistical goodness-of-fit problem into a knowledge-conditioned decision-making problem that integrates observed data with domain knowledge.
\end{abstract}


\section{Introduction}

As the importance of operational stability, reliability, and cost reduction continues to increase, remaining useful life (RUL) prediction has become increasingly crucial across various domains including the aerospace \cite{RN18}, transportation and logistics \cite{RN16}, and energy sectors \cite{RN17}. RUL prediction directly links condition monitoring with operational maintenance decision-making, and is widely regarded as a core component of condition-based maintenance \cite{RN15}. A central task in RUL prediction is to model the future deterioration of health indicators (HIs) collected from, or extracted from, equipment \cite{RN14}. Because inappropriate degradation models can underestimate or overestimate RUL uncertainty, choosing an appropriate degradation model is a critical issue \cite{RN13}.

Despite this need, studies on stochastic process-based degradation modeling have primarily focused on structural improvements of degradation models \cite{RN12} and parameter estimation or approximation methods \cite{RN11}, whereas degradation model selection has received less attention. Existing degradation model selection methods can be broadly divided into likelihood-based and prognostics-oriented criteria based on RUL prediction performance \cite{RN13}. Likelihood-based criteria can enable robust performance when sufficient data are available; however, when the observation window is short or the HI contains substantial noise, they may select a model that does not reflect the actual degradation mechanism \cite{RN5}. Prognostics-oriented criteria can identify degradation trajectories suitable for RUL prediction, but some model selection procedures may implicitly use information that is available only after the prediction time or even the full life-cycle trajectory, thereby causing data leakage. In other words, existing degradation model selection methods rely mainly on the observed HI trajectory; consequently, they are sensitive to data quality and limited in their ability to incorporate knowledge about system-level or equipment-specific failure mechanisms.

In industrial practice, expert knowledge, experience, and technical expertise are recognized as being essential for equipment maintenance \cite{RN10}. This knowledge includes information on the system type, structural characteristics, failure modes, sensor types, and operating conditions, as well as individual tacit knowledge. For example, irreversible damage accumulation processes such as corrosion, cumulative wear, and crack propagation are naturally associated with gamma-family models, which assume non-negative increments and monotonic degradation behavior \cite{RN9}. In contrast, vibration-based HIs in rotating machinery may exhibit repeated local increases and decreases due to rotational frequency variations, load fluctuations, and changes in operating conditions \cite{RN8}. In such cases, Wiener-family models may be more appropriate candidates. Furthermore, even within each degradation family, extended models can be further subdivided according to linearity, the presence of an initial degradation point, change points, and random effects \cite{RN7}. Under these more realistic conditions, degradation model selection should be regarded not merely as a statistical model selection problem, but as a scientific decision-making problem that integrates statistical evidence with an understanding of system behavior and failure mechanisms.

To this end, we propose Literature-Anchored Stochastic Trajectory Retrieval-Augmented Generation (LAST-RAG), a local-evidence-based RAG framework for stochastic degradation model selection. Given an observed HI trajectory and domain-specific context, LAST-RAG retrieves proposition-level evidence from a pre-constructed local evidence bank and uses the retrieved evidence to condition the selection of candidate stochastic degradation models. The performance of the proposed method was validated using simulation-generated datasets. LAST-RAG achieved an average F1 score of 0.925 in binary stochastic-process family classification. Notably, even when only 30\% of the HI trajectory up to the actual failure time was observed, LAST-RAG achieved an F1 score of 0.983, corresponding to an improvement of 0.394 over the state-of-the-art baseline. The main contributions of this study are summarized as follows:

\begin{itemize}

    \item We extend degradation model selection beyond a purely statistics-based problem by formulating it as a knowledge-integrated decision-making problem that incorporates mechanical characteristics and tacit expert knowledge.
    
    \item We propose a degradation model conditioning method based on LAST-RAG, which integrates theoretical knowledge with observed degradation evidence.
    
    \item We develop a hierarchical model selection framework that combines model conditioning with statistical degradation model selection methods.

\end{itemize}

The remainder of this paper is organized as follows: Section 2 reviews previous studies on degradation model selection in the field of prognostics and health management (PHM). Section 3 presents the proposed LAST-RAG-based knowledge-conditioned degradation model selection framework. Section 4 reports a validation of the effectiveness of the proposed method using simulation-generated data. Finally, Section 5 concludes the paper and discusses future research directions.

\section{Related works}

Degradation modeling has been widely adopted because it enables the estimation of the RUL distribution through the first hitting time (FHT) to a predefined failure threshold \cite{RN13}. Among representative stochastic processes, the gamma process assumes monotonically increasing degradation trajectories and is therefore well suited to describing cumulative damage systems. In practice, however, HIs may exhibit repeated local increases and decreases due to sensor noise, variations in operating conditions, or recovery effects; in such cases, the assumption of monotonicity may be inappropriate. In contrast, Wiener-process-based models can represent non-monotonic degradation trajectories through Gaussian increments and a diffusion structure, making them suitable for degradation modeling under local variability. Nevertheless, many RUL estimation studies assume that the underlying degradation process is known in advance, which limits their applicability in environments lacking sufficient domain knowledge and expert experience \cite{RN6}. Therefore, techniques that can select an appropriate degradation model by integrating the observed HI trajectory with limited domain knowledge are required.

In a representative study addressing degradation model selection, Nguyen et al. \cite{RN13} compared model selection criteria for stochastic process models, including Brownian motion, Ornstein–Uhlenbeck processes, and gamma process families, from the perspectives of goodness-of-fit, model complexity, and prognostic relevance. In addition to conventional criteria, such as the Akaike information criterion (AIC), Bayesian information criterion (BIC), minimum description length (MDL), empirical average log-likelihood (EAL), and cross-validation (CV), they also examined RUL-prediction-oriented criteria, including the prognostic horizon criterion (PHC), prognostic accuracy criterion (PAC), and hybrid criterion (HyC). Their results showed that CV tended to favor models with many parameters, whereas complexity-penalized criteria such as AIC, BIC, and MDL tended to favor simpler models with fewer parameters. They also noted that posterior information related to operating conditions was not sufficiently incorporated. Moreover, PHC was found to be weak in identifying the degradation model family, whereas PAC exhibited a paradoxical tendency to select an incorrect model family as the observation period increased.

Yu et al. \cite{RN5} highlighted model selection uncertainty in online RUL prediction and considered both prior parameter uncertainty and degradation model uncertainty. Specifically, they employed a Bayesian-updated expectation conditional maximization algorithm to update uncertain prior parameters and proposed a modified Bayesian model averaging method to represent uncertainty among candidate degradation models using posterior model probabilities. Rather than forcing the selection of a single model at a given time point, this approach sequentially updates the probabilities of candidate models and incorporates them into the RUL estimation.

Zhou et al. \cite{RN4} proposed another composite indicator for degradation model selection. Their study used the number of model parameters as an indicator of model complexity, the mean absolute error (MAE) as an indicator of goodness-of-fit, and L1 normalization as an indicator of generalization. These three indicators were then normalized and linearly transformed, and their weights were estimated based on the RUL prediction performance of historical degradation data. This method can be regarded as an attempt to jointly consider model complexity, prediction accuracy, and generalization performance.

In summary, early studies on degradation model selection mainly considered likelihood-based model complexity, whereas more recent studies have attempted to further improve RUL prediction performance using criteria such as PHC, PAC, HyC, and MAE. However, in real-world settings, degradation model selection does not rely solely on observed data; it is also closely related to physical degradation mechanisms, operating environments, mechanical characteristics, and tacit knowledge derived from field experts’ experience. In particular, when the amount of observed data is limited or the data exhibit high variability, robust model selection using purely data-driven criteria alone is difficult. Existing studies have largely assumed such knowledge implicitly through researchers’ intuition or prior model designs, without sufficiently addressing practical constraints. This study aims to improve the explainability of degradation model selection while additionally accounting for theoretical degradation models, mechanical characteristics, and recursively accumulated empirical knowledge.

\section{Methodology}

This study addresses the problem of selecting a stochastic degradation model using an observed HI trajectory and domain-specific context. The proposed method is a hybrid model selection framework that hierarchically conditions the structural properties of degradation models using LAST-RAG and applies statistical model selection only to the uncertain branch identified by LAST-RAG. To improve the logical validity and reliability of the large language model (LLM) while explicitly extracting uncertain cases, this study introduces Rule-based Confidence Reasoning with Uncertain State (RCRUS). RCRUS refers to a deterministic arbitration rule that first generates an internal LLM answer and a RAG-based LLM answer at each hierarchy, and then determines the final hierarchical decision based on the agreement between the two answers and the difference in their confidence scores. This mechanism is proposed to mitigate excessive model-set conditioning caused by errors in hierarchical conditioning.

In PHM environments, condition-monitoring data and operational information are often security-sensitive \cite{RN3}. In addition, PHM systems frequently have limited computational resources and real-time inference requirements \cite{RN2}, making lightweight designs essential. Accordingly, this study assumes an offline local LLM setting that does not depend on external APIs and proposes a LAST-RAG-based hierarchical model selection framework combined with domain-specific local RAG.

\begin{figure}[htbp]
    \centering
    \includegraphics[width=0.90\textwidth]{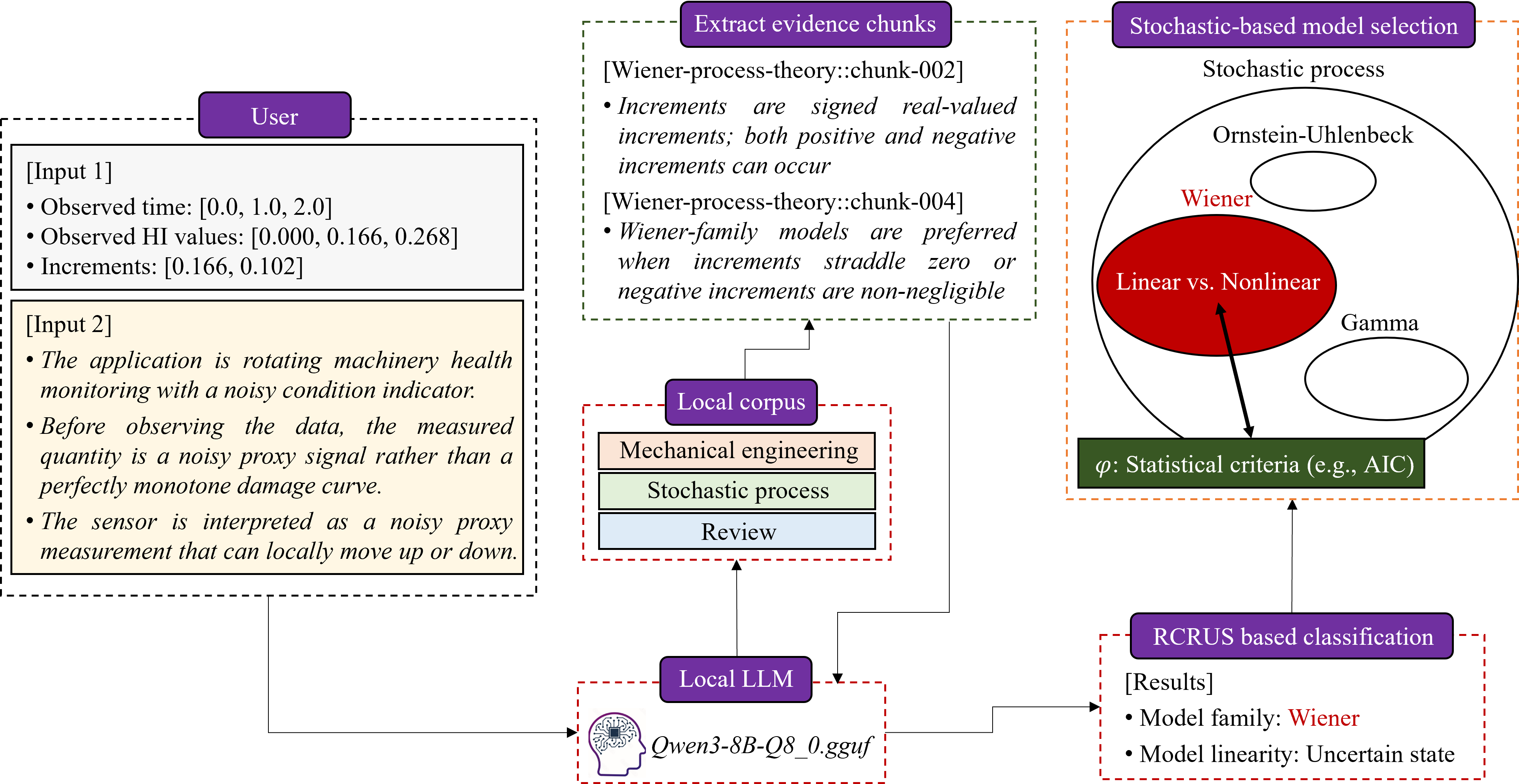}
    \caption{Proposed hierarchical framework combining local LAST-RAG and stochastic degradation model selection.}
    \label{fig:framework}
\end{figure}

\subsection{Problem formulation}
\label{sec:problem-formulation}
For the $i$-th inference scenario, the HI trajectory observed up to inspection time $\tau$ is defined as follows:
\begin{equation}
\mathcal{D}_{\tau}^{i}
= \left\{ \left(t_{ij}, y_{ij}\right) : j = 1,\ldots,n_{\tau}^{i},\; t_{ij} \leq \tau \right\}.
\label{eq:observed-trajectory}
\end{equation}
Here, $t_{ij}$ denotes the $j$-th inspection time in the $i$-th scenario, and $y_{ij}$ denotes the HI value observed at that time. The term $n_{\tau}^{i}$ is the number of HI samples observed up to time $\tau$.

In this study, the operator uses domain context provided by an operator or analyst together with the HI sequence. This is defined as
\begin{equation}
\mathcal{X}_{\tau}^{i}
= \left(\mathcal{D}_{\tau}^{i}, \mathcal{K}_{\tau}^{i}\right),
\label{eq:model-input}
\end{equation}
where $\mathcal{K}_{\tau}^{i}$ denotes mechanical characteristics or experiential knowledge provided in textual form by an operator or analyst. Mechanical characteristics may include, for example, equipment type, the physical meaning of the HI, sensor characteristics, operating environment, wear, cracking, corrosion, fatigue, and battery capacity fading. Experiential knowledge may include experience regarding the variability of the observed HI, past records, and qualitative judgments based on visual factors. However, $\mathcal{X}_{\tau}^{i}$ excludes information that is only available after time $\tau$ and information directly linked to the degradation model selection outcome, such as failure mode, future observations, failure time, and RUL labels. This exclusion reflects realistic constraints in online inference.

The objective of this study is to select the final degradation model $\widehat{m}_{\tau}^{i} \in \mathcal{M}$ from the candidate set of stochastic degradation models using $\mathcal{X}_{\tau}^{i}$:
\begin{equation}
\mathcal{M} = \left\{m_1,m_2,\ldots,m_{|\mathcal{M}|}\right\},
\label{eq:candidate-set}
\end{equation}
where the elements of $\mathcal{M}$ may include, for example, a linear Wiener process or a non-homogeneous gamma process.

\subsection{Hierarchical representation of the candidate model space}
\label{sec:hierarchical-space}
Each candidate model $m \in \mathcal{M}$ is defined as a combination of multiple structural attributes. In this study, these attributes are defined as hierarchies, and the complete set of hierarchies is represented as
\begin{equation}
\mathcal{H} = \left\{F,T\right\},
\label{eq:hierarchies}
\end{equation}
where $F$ denotes the stochastic process family and $T$ denotes the trend structure. Let $\mathcal{Y}_h$ be the label space for each hierarchy $h \in \mathcal{H}$. Then,
\begin{align}
\mathcal{Y}_F &= \left\{W,G\right\}, \label{eq:family-label-space}\\
\mathcal{Y}_T &= \left\{L,NL\right\}, \label{eq:trend-label-space}
\end{align}
where $\mathcal{Y}_F$ is the family condition set, and $W$ and $G$ denote the Wiener and gamma families, respectively. The set $\mathcal{Y}_T$ represents the form of the trend function, where $L$ and $NL$ denote linearity and nonlinearity, respectively.

The structural label assigned to each candidate model $m$ at hierarchy $h$ is defined by the following mapping:
\begin{equation}
\psi_h : \mathcal{M} \rightarrow \mathcal{Y}_h,\qquad \psi_h(m) \in \mathcal{Y}_h.
\label{eq:structural-label-map}
\end{equation}
Accordingly, the complete structural attributes of candidate model $m$ can be expressed as
\begin{equation}
\psi(m) = \left(\psi_h(m)\right)_{h \in \mathcal{H}}.
\label{eq:complete-structural-attributes}
\end{equation}
For example, if a model $m$ belongs to the Wiener family and exhibits a nonlinear trend, it can be represented as
\begin{equation}
\psi(m) = (W,NL).
\label{eq:wiener-nonlinear-example}
\end{equation}

\subsection{Local evidence bank construction and evidence retrieval}
\label{sec:leb}
This study assumes an offline RAG setting, in which all evidence is restricted to a pre-constructed local evidence bank (LEB).

\subsubsection{Offline proposition evidence bank construction}
\label{sec:offline-bank}
The offline literature corpus is defined as
\begin{equation}
\mathcal{P} = \left\{p_l \mid l = 1,\ldots,L\right\},
\label{eq:corpus}
\end{equation}
where $p_l$ denotes the $l$-th document used to construct the LEB. The corpus $\mathcal{P}$ consists of three categories of literature. First, literature on mechanical engineering mechanisms describes physical degradation mechanisms and the meanings of HIs, including wear, corrosion, crack propagation, bearing degradation, and battery capacity fading. Second, stochastic process theory literature provides the mathematical properties of the Wiener and gamma processes. Specifically, Wiener-process-based models are suitable for describing continuous sample paths, Gaussian increments, measurement noise, and local fluctuations. In contrast, gamma-process-based models are suitable for describing non-negative increments, monotonic accumulation, and irreversible degradation. Third, degradation modeling review literature provides theoretical or empirical evidence linking specific degradation patterns to stochastic processes.

Each $p_l$ is decomposed into atomic propositions using an offline proposition tokenizer $\Phi$:
\begin{equation}
Q_l = \Phi(p_l) = \left\{q_{l1},\ldots,q_{lN_l}\right\}.
\label{eq:proposition-tokenizer}
\end{equation}
The role of $\Phi$ is to transform the document into self-contained, minimal, and semantically interpretable proposition-level retrieval units \cite{RN1}. In this study, the publicly available \texttt{propositionizer-wiki-flan-t5-large}\footnote{\url{https://huggingface.co/chentong00/propositionizer-wiki-flan-t5-large}} was used as $\Phi$; further details are provided in \cite{RN1}.\footnote{\url{https://github.com/chentong0/factoid-wiki}} Here, $N_l$ denotes the number of propositions extracted from $p_l$.

The offline LEB is then constructed as follows:
\begin{align}
\ \mathcal{E}^{\mathrm{off}} &= \left\{ e_r = (q_r,s_r) \mid r = 1,\ldots,R \right\}, \label{eq:leb}\\
R &= \sum_{l=1}^L N_l, \label{eq:leb-size}
\end{align}
where $q_r$ denotes the $r$-th proposition, and $s_r$ denotes its provenance metadata, which serve as auxiliary information for tracing the literature source from which the retrieved evidence originates.

\subsubsection{Query construction}
\label{sec:query-construction}
The retrieval query is constructed from $\mathcal{X}_{\tau}^{i}$ as follows:
\begin{equation}
g_{\tau}^{i} = \Omega\left(\mathcal{X}_{\tau}^{i}\right),
\label{eq:query-construction}
\end{equation}
where $\Omega(\cdot)$ denotes a prompt that transforms the HI trajectory and domain-specific context into a textual representation suitable for retrieval. The query $g_{\tau}^{i}$ summarizes the main characteristics of the observed HI trajectory and domain context. For example, a query may include descriptions of monotonicity, local fluctuations, abrupt changes, nonlinear trends, and the possibility of measurement noise. An example of $\Omega$ used in this study is provided in Appendix~\ref{app:prompt-examples}.

\subsubsection{Proposition-level evidence retrieval}
\label{sec:evidence-retrieval}
Based on the generated $g_{\tau}^{i}$, the local LLM retrieves the top-$K$ proposition-level evidence from $\mathcal{E}^{\mathrm{off}}$:
\begin{align}
\mathcal{R}_{\tau}^{i} &= \operatorname{Retrieve}_K\left(g_{\tau}^{i},\mathcal{E}^{\mathrm{off}}\right), \label{eq:retrieve}\\
\mathcal{R}_{\tau}^{i} &= \left\{ e_r = (q_r,s_r) \mid r \in \mathcal{I}_{\tau}^{i} \right\},\qquad
\mathcal{I}_{\tau}^{i} \subseteq \left\{1,\ldots,R\right\}. \label{eq:retrieved-set}
\end{align}
Here, $\mathcal{I}_{\tau}^{i}$ denotes the evidence index set retrieved at inspection time $\tau$ in the $i$-th inference scenario. The retrieved evidence is subsequently used as evidence-scenario input when generating hierarchy-specific LLM decisions.

\subsection{RCRUS: Rule-based Confidence Reasoning with Uncertain State}
\label{sec:rcrus}
In hierarchical model conditioning, an error can condition the entire model set toward a specific subset, which may lead to excessive elimination under weak evidence. RCRUS introduces an uncertain state to mitigate this problem. Specifically, the local LLM generates an answer and a confidence score within the label space of each hierarchy. Then, based on a deterministic arbitration rule, RCRUS evaluates whether the two answers conflict and whether the difference between their confidence scores is sufficiently large, thereby assigning an uncertain state when appropriate.

For each hierarchy $h$, the extended decision space is defined as
\begin{equation}
\widetilde{\mathcal{Y}}_h = \mathcal{Y}_h \cup \left\{U_h\right\},
\label{eq:extended-decision-space}
\end{equation}
where $U_h$ denotes the uncertain state in hierarchy $h$.

First, the answer from the internal local LLM is generated as
\begin{equation}
a_{\mathrm{int},h,\tau}^{i} = \left(z_{\mathrm{int},h,\tau}^{i},c_{\mathrm{int},h,\tau}^{i}\right)
= \Lambda_{\mathrm{int},h}\left(\mathcal{X}_{\tau}^{i}\right),
\label{eq:internal-answer}
\end{equation}
where $\Lambda_{\mathrm{int},h}$ is the internal local LLM decision function for hierarchy $h$. The term $z_{\mathrm{int},h,\tau}^{i}$ denotes the answer label produced by $\Lambda_{\mathrm{int},h}$, and $c_{\mathrm{int},h,\tau}^{i}$ denotes the confidence score of that answer.

Second, using the retrieved evidence $\mathcal{R}_{\tau}^{i}$, the evidence-conditioned answer is generated as
\begin{equation}
a_{\mathrm{ctx},h,\tau}^{i} = \left(z_{\mathrm{ctx},h,\tau}^{i},c_{\mathrm{ctx},h,\tau}^{i}\right)
= \Lambda_{\mathrm{ctx},h}\left(\mathcal{X}_{\tau}^{i},\mathcal{R}_{\tau}^{i}\right),
\label{eq:context-answer}
\end{equation}
where $\Lambda_{\mathrm{ctx},h}$ is the evidence-conditioned local LLM decision function for hierarchy $h$. The term $z_{\mathrm{ctx},h,\tau}^{i}$ denotes the answer label produced by $\Lambda_{\mathrm{ctx},h}$, and $c_{\mathrm{ctx},h,\tau}^{i}$ denotes the corresponding confidence score.

Third, the final hierarchical decision of RCRUS is determined by comparing $z_{\mathrm{int},h,\tau}^{i}$, $c_{\mathrm{int},h,\tau}^{i}$, $z_{\mathrm{ctx},h,\tau}^{i}$, and $c_{\mathrm{ctx},h,\tau}^{i}$. Let the final decision be
\begin{equation}
z_{h,\tau}^{i} \in \widetilde{\mathcal{Y}}_h.
\label{eq:final-hierarchy-decision}
\end{equation}
Then, the RCRUS arbitration rule is defined as follows:
\begin{equation}
\widehat{z}_{h,\tau}^{i}
= \begin{cases}
z_{\mathrm{ctx},h,\tau}^{i}, & \text{if } z_{\mathrm{int},h,\tau}^{i} = z_{\mathrm{ctx},h,\tau}^{i}, \\
z_{\mathrm{int},h,\tau}^{i}, & \text{if } z_{\mathrm{int},h,\tau}^{i} \neq z_{\mathrm{ctx},h,\tau}^{i} \text{ and } c_{\mathrm{int},h,\tau}^{i} > c_{\mathrm{ctx},h,\tau}^{i} + \delta, \\
z_{\mathrm{ctx},h,\tau}^{i}, & \text{if } z_{\mathrm{int},h,\tau}^{i} \neq z_{\mathrm{ctx},h,\tau}^{i} \text{ and } c_{\mathrm{ctx},h,\tau}^{i} > c_{\mathrm{int},h,\tau}^{i} + \delta, \\
U_h, & \text{if } z_{\mathrm{int},h,\tau}^{i} \neq z_{\mathrm{ctx},h,\tau}^{i} \text{ and } \left|c_{\mathrm{int},h,\tau}^{i} - c_{\mathrm{ctx},h,\tau}^{i}\right| \leq \delta.
\end{cases}
\label{eq:rcrus-rule}
\end{equation}
Here, $\delta$ is the confidence margin for the hierarchy. In this study, $\delta=0.05$ is used as the default value for all hierarchies, although it may also be treated as a tunable hyperparameter.

The RCRUS arbitration rule in Eq.~\eqref{eq:rcrus-rule} follows three principles. First, if the internal and evidence-conditioned answers support the same label, that label is selected. Second, if the two answers support different labels, their confidence scores are compared. If one confidence score exceeds the other by at least $\delta$, the answer with the higher confidence score is selected. Third, if the two answers differ but the confidence-score gap is no larger than $\delta$, the uncertain state $U_h$ is returned for that hierarchy, and no conditioning is performed.

\subsection{RCRUS-based hierarchical space conditioning}
\label{sec:space-conditioning}
Once the final decision $\widehat{z}_{h,\tau}^{i}$ for each hierarchy is determined by RCRUS, the candidate model space is conditioned accordingly.

First, the candidate subspace corresponding to label $y \in \mathcal{Y}_h$ in hierarchy $h$ is defined as
\begin{equation}
\mathcal{M}_h(y) = \left\{m \in \mathcal{M} \mid \psi_h(m) = y\right\}.
\label{eq:candidate-subspace}
\end{equation}
If $\widehat{z}_{h,\tau}^{i}=y$, only models belonging to $\mathcal{M}_h(y)$ are retained in hierarchy $h$. In contrast, for $\widehat{z}_{h,\tau}^{i}=U_h$, no conditioning is performed. Therefore, the retained candidate model set at time $\tau$ is defined as
\begin{equation}
\mathcal{M}_{\tau}^{i}
= \left\{m \in \mathcal{M} \mid \forall h \in \mathcal{H},\; \widehat{z}_{h,\tau}^{i}=U_h \;\text{or}\; \psi_h(m)=\widehat{z}_{h,\tau}^{i} \right\}.
\label{eq:retained-set}
\end{equation}
Equivalently, if the set of hierarchies for which confident decisions are generated is defined as
\begin{equation}
\mathcal{H}_{\mathrm{conf},\tau}^{i} = \left\{h \in \mathcal{H} \mid \widehat{z}_{h,\tau}^{i} \neq U_h\right\},
\label{eq:confident-hierarchies}
\end{equation}
then
\begin{equation}
\mathcal{M}_{\tau}^{i}
= \bigcap_{h \in \mathcal{H}_{\mathrm{conf},\tau}^{i}} \mathcal{M}_h\left(\widehat{z}_{h,\tau}^{i}\right),
\label{eq:intersection-retained-set}
\end{equation}
with the convention that the intersection equals $\mathcal{M}$ when $\mathcal{H}_{\mathrm{conf},\tau}^{i}$ is empty.

For example, if the hierarchy consists of the family and trend, and the RCRUS decisions are
\begin{equation}
\widehat{z}_{F,\tau}^{i} = W \quad \text{and} \quad \widehat{z}_{T,\tau}^{i} = L,
\label{eq:example-confident-decisions}
\end{equation}
then the retained candidate set is
\begin{equation}
\mathcal{M}_{\tau}^{i}
= \mathcal{M}_F(W) \cap \mathcal{M}_T(L)
= \left\{m \in \mathcal{M} \mid \psi_F(m)=W,\; \psi_T(m)=L\right\}.
\label{eq:example-confident-retained-set}
\end{equation}
As another example, if the RCRUS decisions include an uncertain state, such as
\begin{equation}
\widehat{z}_{F,\tau}^{i} = W \quad \text{and} \quad \widehat{z}_{T,\tau}^{i} = U_T,
\label{eq:example-uncertain-decisions}
\end{equation}
then the retained candidate set is
\begin{equation}
\mathcal{M}_{\tau}^{i}
= \mathcal{M}_F(W)
= \left\{m \in \mathcal{M} \mid \psi_F(m)=W\right\}.
\label{eq:example-uncertain-retained-set}
\end{equation}
In this case, because the trend hierarchy is assigned the uncertain state, $\widehat{z}_{T,\tau}^{i}$ is determined in the subsequent statistical model selection stage.

\subsection{Stochastic model selection within the retained candidate set}
\label{sec:stochastic-selection}
After RCRUS-based hierarchical space conditioning, if the retained candidate set $\mathcal{M}_{\tau}^{i}$ contains only a single model, that model is selected as the final degradation model. Formally,
\begin{equation}
\widehat{m}_{\tau}^{i} = m, \qquad \text{if } \mathcal{M}_{\tau}^{i} = \left\{m\right\}.
\label{eq:single-candidate-selection}
\end{equation}
In contrast, if multiple candidate models remain because at least one uncertain state $U_h$ exists, statistical model selection is performed based on the observed HI trajectory $\mathcal{D}_{\tau}^{i}$. Let $S(m;\mathcal{D}_{\tau}^{i})$ denote the model selection score for candidate model $m$, where a smaller value of $S(\cdot)$ indicates a more suitable model. The final selected model is then determined as
\begin{equation}
\widehat{m}_{\tau}^{i}
= \arg\min_{m \in \mathcal{M}_{\tau}^{i}} S\left(m;\mathcal{D}_{\tau}^{i}\right).
\label{eq:score-based-selection}
\end{equation}
In this study, EAL is used as $S(\cdot)$; however, the proposed framework is not dependent on a specific model selection criterion.

\section{Experiments}
\label{sec:experiments}

\subsection{Dataset}
\label{sec:dataset}
Because it is difficult to obtain ground-truth labels for specific degradation curves, this study defined conventional stochastic process models and used datasets generated through simulations for validation. The experiments were divided into Cases 1 and 2 according to the task difficulty, as summarized in Table~\ref{tab:cases}. Case 1 addresses binary classification at the gamma- and Wiener-family level, whereas Case 2 considers a four-class classification task that distinguishes not only the gamma and Wiener families but also linear/nonlinear trends and homogeneity/non-homogeneity.

\begin{table}[htbp]
\small
\centering
\caption{Stochastic process models included in each experimental case.}
\label{tab:cases}
\footnotesize
\begin{tabular}{llll}
\toprule
Experiment case & Family & Degradation model & Problem \\
\midrule
Case 1 & Wiener & Wiener family & 2-class classification \\
       & Gamma  & Gamma family  & \\
\midrule
Case 2 & Wiener & Linear Wiener & 4-class classification \\
       &        & Nonlinear Wiener & \\
       & Gamma  & Homogeneous gamma & \\
       &        & Non-homogeneous gamma & \\
\bottomrule
\end{tabular}
\end{table}

\subsection{Experimental settings}
\label{sec:experimental-settings}
For each stochastic degradation model described in Section~\ref{sec:dataset}, independent run-to-failure (RtF) degradation trajectories were generated using the simulation parameters summarized in Appendix~\ref{app:implementation-details}. Each trajectory corresponded to one independent unit, and no trajectory shared noise realizations or random seeds with any other trajectory. The generated trajectories were divided into training, validation, and test sets at a ratio of 60\%, 20\%, and 20\%, respectively. Splitting was performed at the trajectory level.

The training and validation sets were used for parameter estimation and model fitting, whereas the final performance was exclusively evaluated on the test set. All comparative methods were evaluated using the same training, validation, and test splits and the same set of candidate models.

\begin{figure}[htbp]
    \centering
    \includegraphics[width=0.90\textwidth]{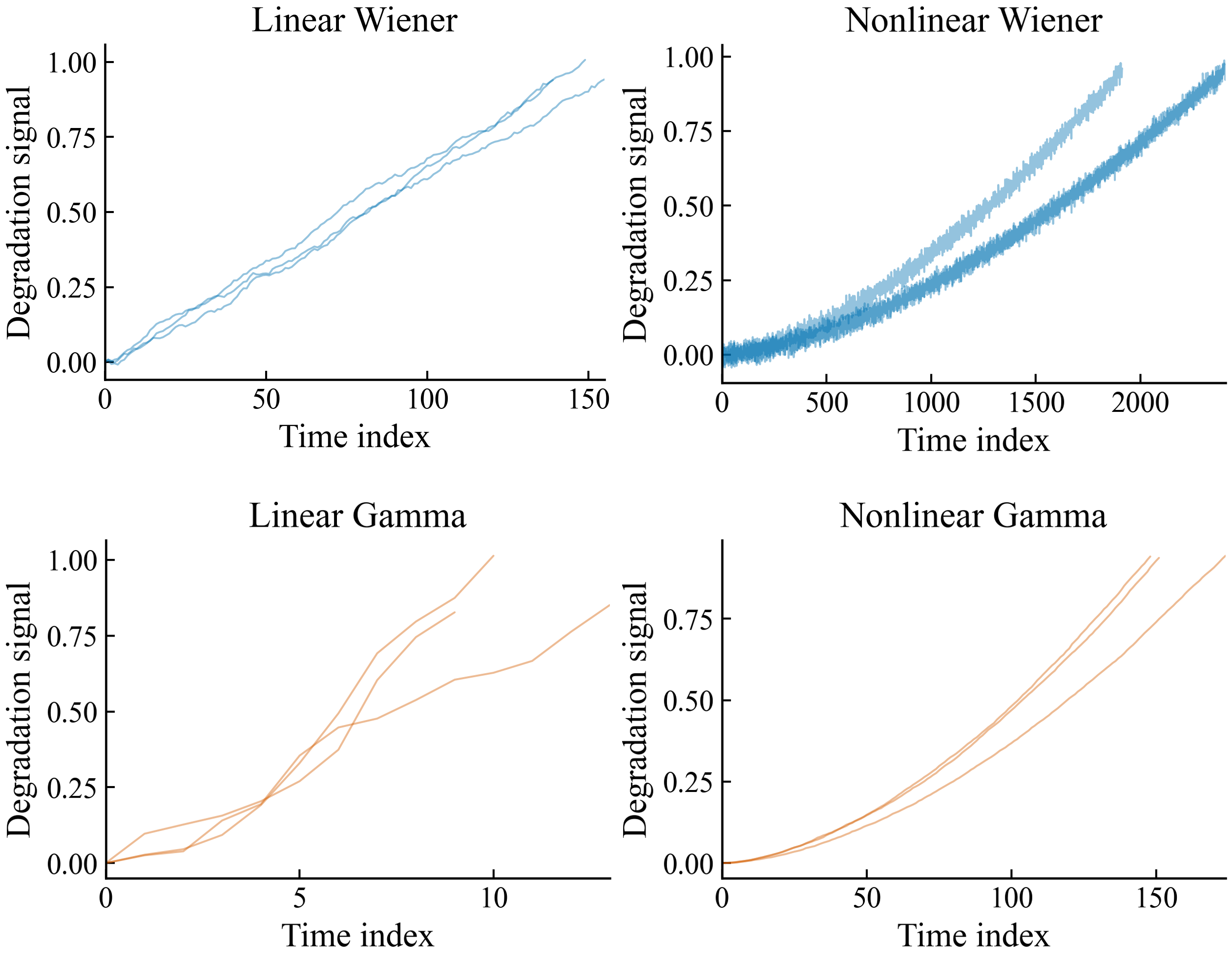}
    \caption{Four models in the Wiener- and gamma-process family.}
    \label{fig:processes}
\end{figure}

To examine model selection performance under different observation horizons, each RtF trajectory was truncated according to the degradation progression rate $n$ (\%), where $n \in \left\{30,50,70\right\}$. The stochastic processes considered in this study are illustrated in Figure~\ref{fig:processes}.

\begin{table}[htbp]
\centering
\small
\caption{Baselines for degradation model selection.}
\label{tab:baselines}
\begin{tabular}{lll}
\toprule
Category & Author & Selection criterion \\
\midrule
Statistical & -- & AIC, BIC, MDL, CV, EAL \\
Prognostic & Nguyen et al. & PHC, PAC, HyC \\
            & Zhuo et al. & LC \\
Uncertainty & Yu et al. & BAM \\
\bottomrule
\end{tabular}
\end{table}

The performance of the proposed method was compared with that of three groups of model selection baselines, as summarized in Table~\ref{tab:baselines}. The first group consisted of statistical criteria including AIC, BIC, MDL, CV, and EAL. The second group comprised prognostic criteria including PHC, PAC, HyC, and LC. The third group was the Bayesian averaging model (BAM), which considers uncertainty in model selection. Classification performance was measured using accuracy (Acc), precision (Prec), recall (Rec), and F1-score (F1).

\subsection{Experimental results}
\label{sec:experimental-results}
This section reports two experimental results. First, the classification performance of the degradation models is evaluated for Cases 1 and 2. Second, the robustness of the proposed family classification framework is examined under information perturbation in the input data.

\begin{table}[H]
\centering
\small
\caption{Classification performance comparison of degradation model selection methods under different observation lengths in Cases 1 and 2. Bold values denote the proposed method; underlined values denote the strongest baseline(s) by F1-score within each observation-length block.}
\label{tab:classification}
\resizebox{\textwidth}{!}{%
\begin{tabular}{llrrrr@{\hspace{0.5cm}}llrrrr}
\toprule
\multicolumn{6}{c}{Case 1: family classification (sample: 24)} & \multicolumn{6}{c}{Case 2: detailed classification (sample: 96)} \\
\cmidrule(r){1-6}\cmidrule(l){7-12}
$n$ & Model & Acc & Prec & Rec & F1 & $n$ & Model & Acc & Prec & Rec & F1 \\
\midrule
30 & AIC & 0.478 & 0.244 & 0.478 & 0.323 & 30 & AIC & 0.375 & 0.594 & 0.375 & 0.333 \\
 & BIC & 0.478 & 0.244 & 0.478 & 0.323 &  & BIC & 0.250 & 0.088 & 0.250 & 0.130 \\
 & MDL & 0.478 & 0.244 & 0.478 & 0.323 &  & MDL & 0.250 & 0.088 & 0.250 & 0.130 \\
 & \secondbest{CV} & \secondbest{0.597} & \secondbest{0.604} & \secondbest{0.597} & \secondbest{0.589} &  & CV & 0.250 & 0.155 & 0.250 & 0.183 \\
 & EAL & 0.478 & 0.244 & 0.478 & 0.323 &  & EAL & 0.500 & 0.303 & 0.500 & 0.361 \\
 & PHC & 0.500 & 0.250 & 0.500 & 0.333 &  & PHC & 0.250 & 0.079 & 0.250 & 0.120 \\
 & PAC & 0.490 & 0.485 & 0.490 & 0.443 &  & PAC & 0.333 & 0.250 & 0.333 & 0.286 \\
 & HyC & 0.471 & 0.288 & 0.471 & 0.325 &  & \secondbest{HyC} & \secondbest{0.625} & \secondbest{0.562} & \secondbest{0.625} & \secondbest{0.548} \\
 & LC & 0.531 & 0.686 & 0.531 & 0.408 &  & LC & 0.250 & 0.083 & 0.250 & 0.125 \\
 & BAM & 0.478 & 0.244 & 0.478 & 0.323 &  & BAM & 0.500 & 0.303 & 0.500 & 0.361 \\
 & \textbf{Proposed} & \textbf{0.968} & \textbf{1.000} & \textbf{0.968} & \textbf{0.983} &  & \textbf{Proposed} & \textbf{0.750} & \textbf{0.667} & \textbf{0.667} & \textbf{0.667} \\
\midrule
50 & AIC & 0.500 & 0.250 & 0.500 & 0.333 & 50 & AIC & 0.250 & 0.088 & 0.250 & 0.130 \\
 & BIC & 0.500 & 0.250 & 0.500 & 0.333 &  & BIC & 0.250 & 0.088 & 0.250 & 0.130 \\
 & MDL & 0.500 & 0.250 & 0.500 & 0.333 &  & MDL & 0.250 & 0.088 & 0.250 & 0.130 \\
 & \secondbest{CV} & \secondbest{0.597} & \secondbest{0.608} & \secondbest{0.597} & \secondbest{0.586} &  & CV & 0.167 & 0.067 & 0.167 & 0.095 \\
 & EAL & 0.500 & 0.250 & 0.500 & 0.333 &  & EAL & 0.500 & 0.303 & 0.500 & 0.361 \\
 & PHC & 0.500 & 0.250 & 0.500 & 0.333 &  & PHC & 0.250 & 0.079 & 0.250 & 0.120 \\
 & PAC & 0.484 & 0.476 & 0.484 & 0.436 &  & PAC & 0.292 & 0.250 & 0.292 & 0.269 \\
 & HyC & 0.487 & 0.246 & 0.487 & 0.327 &  & \secondbest{HyC} & \secondbest{0.542} & \secondbest{0.562} & \secondbest{0.542} & \secondbest{0.452} \\
 & LC & 0.531 & 0.758 & 0.531 & 0.399 &  & LC & 0.250 & 0.094 & 0.250 & 0.136 \\
 & BAM & 0.500 & 0.250 & 0.500 & 0.333 &  & BAM & 0.500 & 0.303 & 0.500 & 0.361 \\
 & \textbf{Proposed} & \textbf{0.906} & \textbf{1.000} & \textbf{0.906} & \textbf{0.950} &  & \textbf{Proposed} & \textbf{0.875} & \textbf{0.688} & \textbf{0.750} & \textbf{0.714} \\
\midrule
70 & AIC & 0.500 & 0.250 & 0.500 & 0.333 & 70 & AIC & 0.250 & 0.083 & 0.250 & 0.125 \\
 & BIC & 0.500 & 0.250 & 0.500 & 0.333 &  & BIC & 0.250 & 0.083 & 0.250 & 0.125 \\
 & MDL & 0.500 & 0.250 & 0.500 & 0.333 &  & MDL & 0.250 & 0.083 & 0.250 & 0.125 \\
 & \secondbest{CV} & \secondbest{0.609} & \secondbest{0.628} & \secondbest{0.609} & \secondbest{0.595} &  & CV & 0.208 & 0.078 & 0.208 & 0.114 \\
 & EAL & 0.500 & 0.250 & 0.500 & 0.333 &  & \secondbest{EAL} & \secondbest{0.500} & \secondbest{0.333} & \secondbest{0.500} & \secondbest{0.375} \\
 & PHC & 0.500 & 0.250 & 0.500 & 0.333 &  & PHC & 0.250 & 0.079 & 0.250 & 0.120 \\
 & PAC & 0.487 & 0.485 & 0.487 & 0.468 &  & PAC & 0.292 & 0.188 & 0.292 & 0.217 \\
 & HyC & 0.500 & 0.250 & 0.500 & 0.333 &  & \secondbest{HyC} & \secondbest{0.500} & \secondbest{0.333} & \secondbest{0.500} & \secondbest{0.375} \\
 & LC & 0.525 & 0.706 & 0.525 & 0.391 &  & LC & 0.250 & 0.083 & 0.250 & 0.125 \\
 & BAM & 0.500 & 0.250 & 0.500 & 0.333 &  & \secondbest{BAM} & \secondbest{0.500} & \secondbest{0.333} & \secondbest{0.500} & \secondbest{0.375} \\
 & \textbf{Proposed} & \textbf{0.750} & \textbf{1.000} & \textbf{0.750} & \textbf{0.843} &  & \textbf{Proposed} & \textbf{0.625} & \textbf{0.667} & \textbf{0.750} & \textbf{0.625} \\
\bottomrule
\end{tabular}%
}
\end{table}

Table~\ref{tab:classification} summarizes the degradation model classification results for Cases 1 and 2 with different observation lengths. Overall, the baseline methods showed case-dependent and unstable performance. In Case 1, CV achieved the strongest performance among the statistical selection criteria, while PAC showed relatively competitive results among the prognostics-oriented criteria. However, their F1-scores remained limited, indicating insufficient discriminative capability even in the binary classification setting. In Case 2, the best-performing baseline varied across $n$, with HyC, EAL, and BAM showing better results, but no baseline consistently maintained high performance.

By contrast, the proposed method achieved the best overall performance in both cases. In Case 1, its F1-scores were 0.983, 0.950, and 0.843 for $n=30$, 50, and 70, respectively. In Case 2, it achieved F1-scores of 0.667, 0.714, and 0.625, outperforming the strongest baseline in all settings. These results indicate that the proposed method provides more reliable degradation model classification than the conventional information criteria, likelihood-based criteria, and prognostics-oriented selection measures. The consistent improvement across cases suggests that the proposed framework can effectively exploit both trajectory-level degradation patterns and model-relevant evidence for robust model selection.

\begin{table}[H]
\centering
\small
\caption{Robustness analysis of stochastic-process family classification under HI trajectory and domain-specific perturbations.}
\label{tab:robustness}
\begin{tabular}{lllrrrr}
\toprule
$n$ & HI trajectory & Domain context & Acc & Prec & Rec & F1 \\
\midrule
30 & \textbf{Correct} & \textbf{Correct} & \textbf{0.875} & 1.000 & \textbf{0.875} & \textbf{0.929} \\
   & \secondbest{Correct} & \secondbest{Wrong} & \secondbest{0.750} & 1.000 & \secondbest{0.750} & \secondbest{0.833} \\
   & Wrong & Correct & 0.000 & 0.000 & 0.000 & 0.000 \\
   & Wrong & Wrong & 0.000 & 0.000 & 0.000 & 0.000 \\
\midrule
50 & \textbf{Correct} & \textbf{Correct} & \textbf{1.000} & 1.000 & \textbf{1.000} & \textbf{1.000} \\
   & \secondbest{Correct} & \secondbest{Wrong} & \secondbest{0.750} & 1.000 & \secondbest{0.750} & \secondbest{0.833} \\
   & Wrong & Correct & 0.000 & 0.000 & 0.000 & 0.000 \\
   & Wrong & Wrong & 0.000 & 0.000 & 0.000 & 0.000 \\
\midrule
70 & \textbf{Correct} & \textbf{Correct} & \textbf{1.000} & 1.000 & \textbf{1.000} & \textbf{1.000} \\
   & \secondbest{Correct} & \secondbest{Wrong} & \secondbest{0.750} & 1.000 & \secondbest{0.750} & \secondbest{0.833} \\
   & Wrong & Correct & 0.000 & 0.000 & 0.000 & 0.000 \\
   & Wrong & Wrong & 0.000 & 0.000 & 0.000 & 0.000 \\
\bottomrule
\end{tabular}
\end{table}

Table~\ref{tab:robustness} presents an evaluation of the effect of information perturbation in the HI trajectory and domain context on performance at the family classification stage. When both types of information were correctly provided, the proposed method achieved an F1-score of 0.929 at $n=30$, and all performance metrics reached 1.000 at $n=50$ and $n=70$. When only the domain context was perturbed, accuracy and F1-score decreased to 0.750 and 0.833, respectively, but the classification performance remained high. This suggests that the proposed method utilizes domain context without excessively relying on it. In contrast, when the HI trajectory was perturbed, all performance metrics decreased to 0.000, regardless of the correctness of the domain context. These results indicate that the family classification decision of the proposed method depends more strongly on the dynamic characteristics of the observed HI trajectory than on the domain context. Therefore, ensuring the quality and reliability of HI trajectories is a critical prerequisite for practical applications.

\section{Conclusion}
\label{sec:conclusion}
This study redefined stochastic degradation model selection in PHM as a knowledge-conditioned decision-making problem that jointly considers observed data and domain knowledge. The proposed LAST-RAG framework hierarchically conditions the stochastic process family and trend structure by leveraging literature-grounded evidence retrieved from a local evidence bank, while RCRUS prevents excessive elimination of candidate models when decisions are uncertain. The experimental results demonstrate that the proposed method achieves more stable performance than existing model selection criteria, even under limited observation windows, suggesting that knowledge-based model conditioning is effective for online RUL inference. Nevertheless, this study was based on simulated degradation trajectories. Future work should validate the generalizability of the proposed framework using real-world PHM datasets and extend it to multivariate HIs, random effects, change-point structures, and covariate-dependent degradation models.
\clearpage
\appendix

\section{Implementation Details for Stochastic Process Generation}
\label{app:implementation-details}
This appendix describes the complete data-generation process, including the distributional assumptions for increments and the definition of change points.

This study considers a four-class classification problem consisting of the linear Wiener process, nonlinear Wiener process, homogeneous gamma process, and non-homogeneous gamma process. For each unit $i$, the degradation state is denoted by $X_{ij}$, where the observation time step is given by $j=0,1,\ldots,N_i$. The initial state is fixed as $X_{i0}=0$, and the increment is defined as $\Delta X_{ij}=X_{ij}-X_{i,j-1}$.

\subsection{Wiener process family}
The Wiener process is a stochastic degradation process with Gaussian increments. The normal distribution is denoted by $N(\mu,\sigma^2)$, where $\mu$ and $\sigma^2$ denote the mean and variance, respectively.

The linear Wiener process represents the case in which both the degradation rate and noise scale remain constant over time, while negative degradation increments are allowed. Its increment is defined as
\begin{equation}
\Delta X_{ij} \sim N(m,s^2),
\label{eq:linear-wiener-increment}
\end{equation}
where $m$ is the constant drift parameter and $s$ is the diffusion parameter.

The nonlinear Wiener process represents the case in which the degradation rate varies over time while negative increments are still allowed. In this study, nonlinearity is introduced using the time-transformed increment function
\begin{equation}
d_j(\beta)=j^\beta-(j-1)^\beta.
\label{eq:time-transformed-increment}
\end{equation}
Accordingly, the increment of the nonlinear Wiener process is defined as
\begin{equation}
\Delta X_{ij} \sim N\left(m d_j(\beta),s^2\right).
\label{eq:nonlinear-wiener-increment}
\end{equation}

\subsection{Gamma process family}
The gamma process is a degradation process whose increments follow a gamma distribution; it is characterized by cumulative and monotonic degradation behavior. The gamma distribution is denoted by $\Gamma(k,\vartheta)$, where $k$ is the shape parameter and $\vartheta$ is the scale parameter.

The homogeneous gamma process represents the case in which the non-negative degradation increments follow a time-invariant structure. Its increment is defined as
\begin{equation}
\Delta X_{ij} \sim \Gamma(\alpha,\vartheta).
\label{eq:homogeneous-gamma-increment}
\end{equation}
The non-homogeneous gamma process represents the case in which the rate of non-negative degradation varies over time. As in the nonlinear Wiener process, the time-transformed increment function $d_j(\beta)=j^\beta-(j-1)^\beta$ is used to represent nonlinearity. Accordingly, the increment of the non-homogeneous gamma process is defined as
\begin{equation}
\Delta X_{ij} \sim \Gamma\left(\alpha d_j(\beta),\vartheta\right).
\label{eq:nonhomogeneous-gamma-increment}
\end{equation}

\clearpage
\section{Prompt examples}
\label{app:prompt-examples}
\begin{figure}[htbp]
    \centering
    \includegraphics[width=0.90\textwidth]{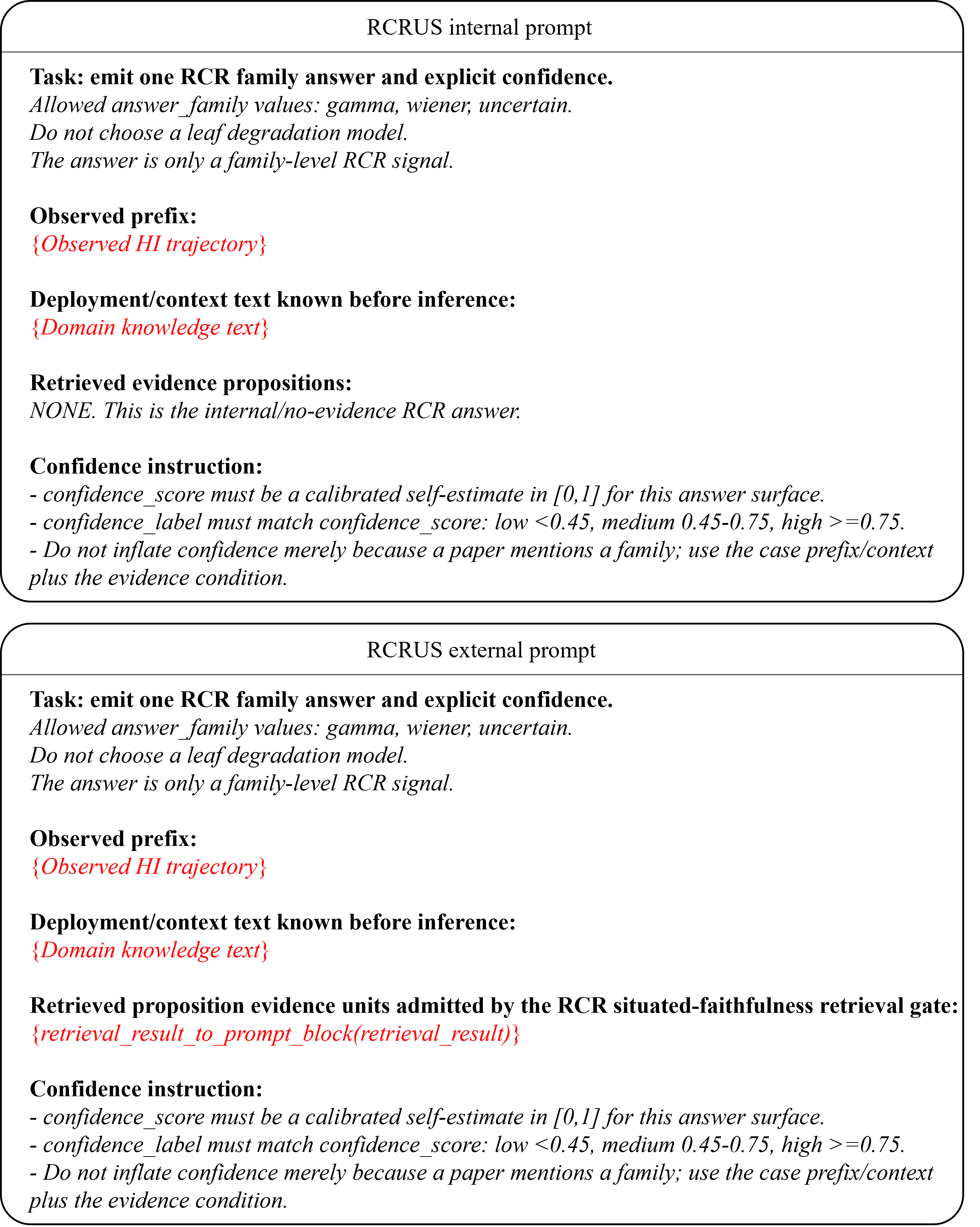}
    \caption{Internal and external LLM prompts used in RCRUS. Bold text indicates the main topic, and red text denotes variables.}
    \label{fig:prompt-examples}
\end{figure}

\clearpage
\section{Domain-specific machinery descriptions}
\label{app:machinery-descriptions}
\begin{figure}[htbp]
    \centering
    \includegraphics[width=0.90\textwidth]{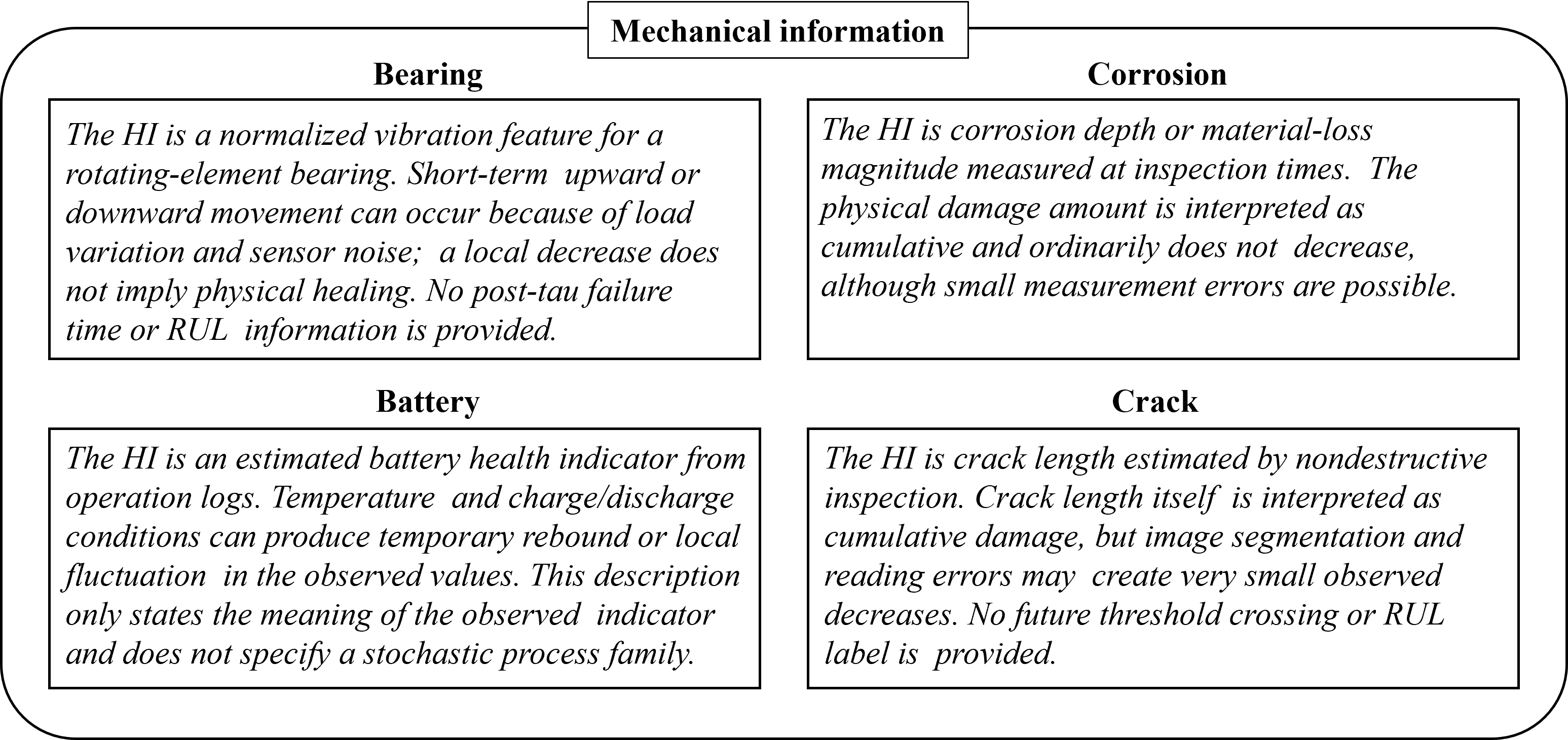}
    \caption{Results retrieved by the local LLM for four types of mechanical information.}
    \label{fig:machinery-descriptions}
\end{figure}

\section{Example evidence chunks extracted from the LEB}
\label{app:evidence-chunks}
\begin{figure}[htbp]
    \centering
    \includegraphics[width=0.90\textwidth]{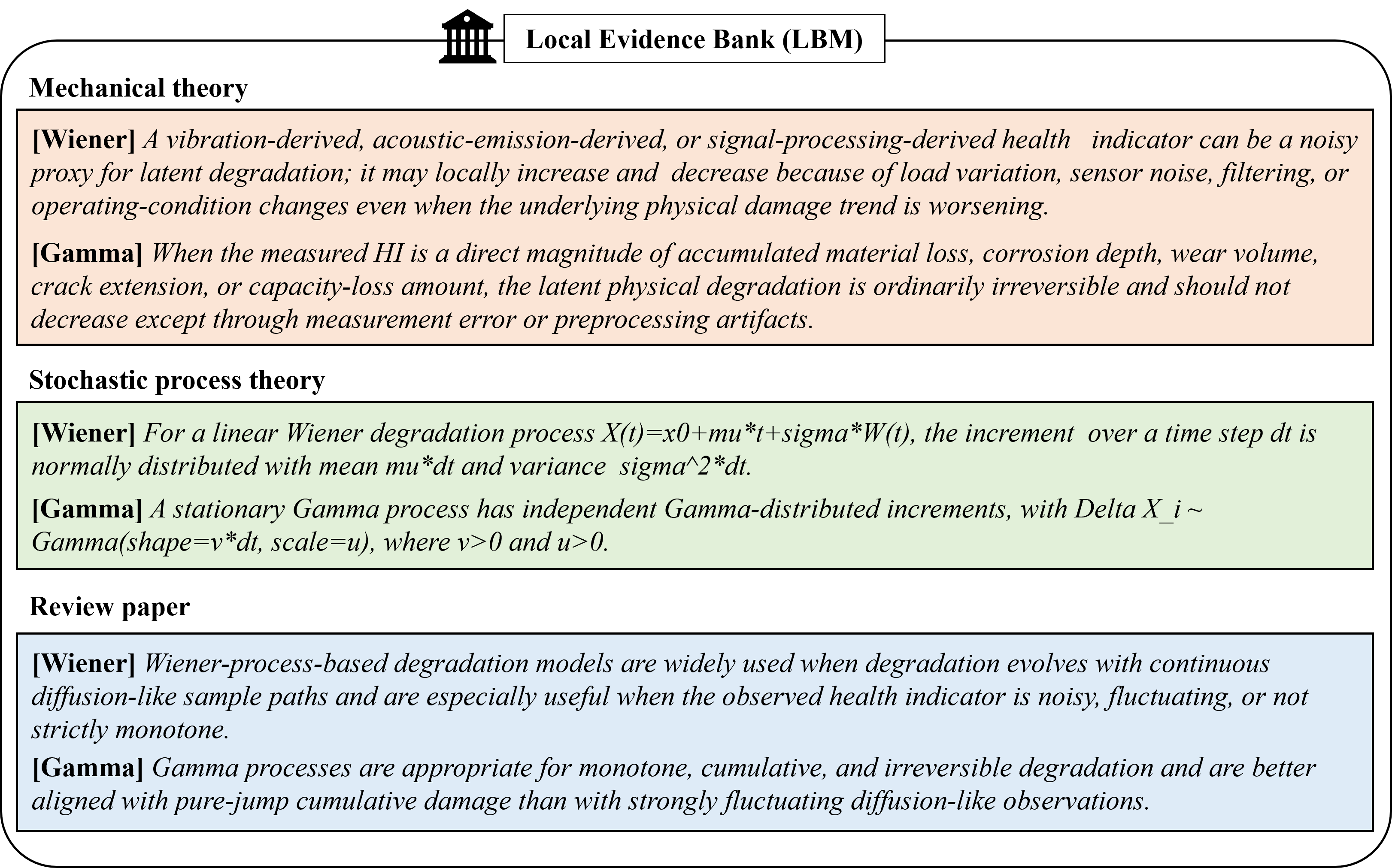}
    \caption{Examples of LEB evidence chunks consisting of mechanical theory, stochastic process theory, and review papers. For each corpus, one representative example corresponding to the Wiener process and one corresponding to the gamma process is presented.}
    \label{fig:evidence-chunks}
\end{figure}

\bibliographystyle{unsrt}
\bibliography{references}

\end{document}